%% file: main.tex
\documentclass[10pt,twocolumn,letterpaper]{article}

\usepackage{cvpr}              


\definecolor{cvprblue}{rgb}{0.21,0.49,0.74}
\usepackage[pagebackref]{hyperref}

\usepackage{booktabs}
\usepackage{multirow}
\usepackage{amsfonts,amssymb,amsmath}
\usepackage{xcolor}         
\usepackage{colortbl}
\usepackage{float}
\usepackage{caption}
\usepackage{graphicx}
\usepackage{natbib}  
\hypersetup{colorlinks}

\def\paperID{8715} 
\def\confName{CVPR}
\def\confYear{2025}

\title{MonoSplat: Generalizable 3D Gaussian Splatting from \\ Monocular Depth Foundation Models}

\author{
    Yifan Liu\textsuperscript{1}\quad
    Keyu Fan\textsuperscript{2}\quad
    Weihao Yu\textsuperscript{1}\quad
    Chenxin Li\textsuperscript{1}\quad
    Hao Lu\textsuperscript{3}\quad
    Yixuan Yuan\textsuperscript{1,}\thanks{Yixuan Yuan is the corresponding author.}\\
    \textsuperscript{1}\small The Chinese University of Hong Kong \quad
    \textsuperscript{2}\small Tsinghua University (Shenzhen) \quad
    \textsuperscript{3}\small HKUST (Guangzhou)\\
    {\tt\footnotesize \{yfliu,yuweihao,chenxinli\}@link.cuhk.edu.hk \quad yxyuan@ee.cuhk.edu.hk}
}

\input{math_commands}

\begin{document}
\maketitle
\input{sec/0_abstract}    
\input{sec/1_intro}
\input{sec/2_related}

\input{sec/3_method}
\input{sec/4_experiments}

\input{sec/5_conclusion}
\input{sec/6_acknowledgement}

{
    \small
    \bibliographystyle{ieeenat_fullname}
    \bibliography{main}
}

\input{sec/X_suppl}

\end{document}

%% file: math_commands.tex
\usepackage{amsmath,amsfonts,bm}

\def\eqref#1{equation~\ref{#1}}

\def\1{\bm{1}}

\def\vc{{\bm{c}}}

\def\vt{{\bm{t}}}

\def\mC{{\bm{C}}}
\def\mD{{\bm{D}}}

\def\mF{{\bm{F}}}
\def\mG{{\bm{G}}}

\def\mI{{\bm{I}}}

\def\mK{{\bm{K}}}

\def\mP{{\bm{P}}}

\def\mR{{\bm{R}}}

\DeclareMathAlphabet{\mathsfit}{\encodingdefault}{\sfdefault}{m}{sl}
\SetMathAlphabet{\mathsfit}{bold}{\encodingdefault}{\sfdefault}{bx}{n}

\def\gI{{\mathcal{I}}}

\def\gP{{\mathcal{P}}}

\newcommand{\R}{\mathbb{R}}



%% file: sec/0_abstract.tex
\begin{abstract}
Recent advances in generalizable 3D Gaussian Splatting have demonstrated promising results in real-time high-fidelity rendering without per-scene optimization, yet existing approaches still struggle to handle unfamiliar visual content during inference on novel scenes due to limited generalizability. To address this challenge, we introduce MonoSplat, a novel framework that leverages rich visual priors from pre-trained monocular depth foundation models for robust Gaussian reconstruction. Our approach consists of two key components: a Mono-Multi Feature Adapter that transforms monocular features into multi-view representations, coupled with an Integrated Gaussian Prediction module that effectively fuses both feature types for precise Gaussian generation. Through the Adapter's lightweight attention mechanism, features are seamlessly aligned and aggregated across views while preserving valuable monocular priors, enabling the Prediction module to generate Gaussian primitives with accurate geometry and appearance. Through extensive experiments on diverse real-world datasets, we convincingly demonstrate that MonoSplat achieves superior reconstruction quality and generalization capability compared to existing methods while maintaining computational efficiency with minimal trainable parameters. Codes are available at \url{https://github.com/CUHK-AIM-Group/MonoSplat}.

\end{abstract}

%% file: sec/1_intro.tex
\section{Introduction}
\label{sec:intro}
The ability to faithfully reconstruct 3D scenes and generate photorealistic renderings represents a cornerstone challenge in computer vision and graphics, with far-reaching implications for virtual reality experiences~\cite{xu2023vr}, digital content creation~\cite{poole2022dreamfusion}, and large-scale visualization systems~\cite{tancik2022block}. While the field has witnessed significant evolution in scene representation approaches - progressing from scene representation networks (SRN)~\cite{sitzmann2019scene} to neural radiance fields (NeRF)~\cite{mildenhall2020nerf}, light field networks (LFN)~\cite{sitzmann2021light}, and most recently 3D Gaussian Splatting (3DGS)~\cite{kerbl20233d} - practical deployment still faces critical barriers. These include substantial computational overhead during rendering~\cite{wang2021ibrnet,yu2021pixelnerf} and time-consuming per-scene optimization procedures~\cite{niemeyer2022regnerf,zhu2023fsgs}, which collectively hinder the widespread adoption of these promising technologies.

\begin{figure}[t!]
  \centering
  \includegraphics[width=0.9\linewidth]{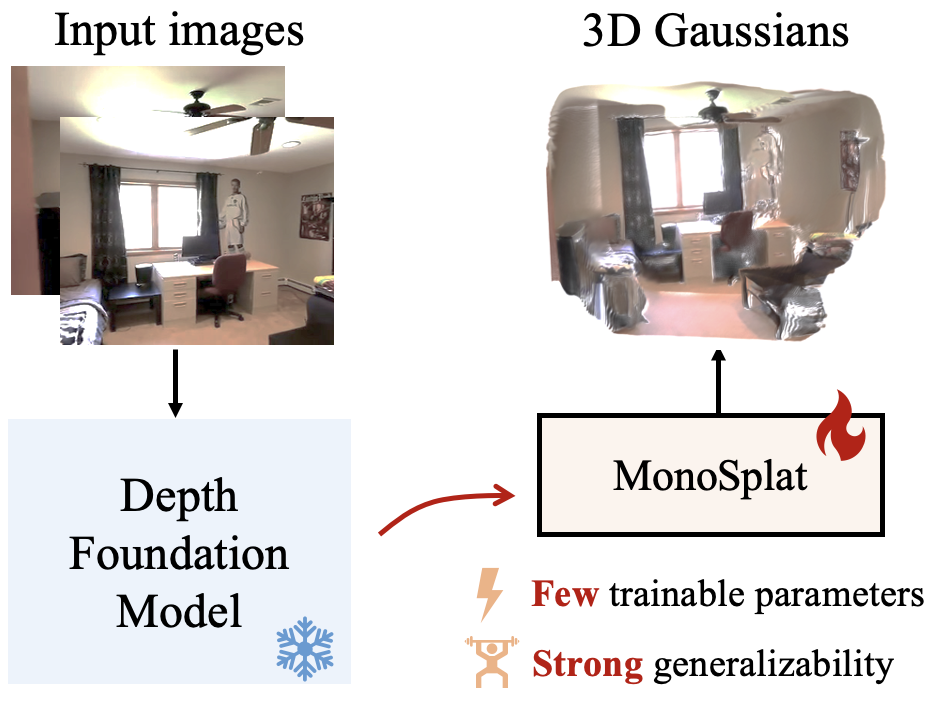}
   \caption{\textbf{Sketch of the proposed MonoSplat framework.} We repurpose the depth foundation model to generalizable Gaussian reconstruction framework.}
   \label{fig:framework}
\end{figure}

Recent advances in generalizable 3D Gaussian Splatting~\cite{charatan2023pixelsplat,chen2024mvsplat,wewer2024latentsplat,liu2024fast} represent a paradigm shift in addressing traditional reconstruction challenges. These methods leverage feed-forward networks for direct 3D Gaussian prediction, allowing for real-time high-fidelity rendering while eliminating the need for per-scene optimization. Various architectural innovations have emerged, including epipolar geometry-based depth estimation in pixelSplat~\cite{charatan2023pixelsplat} and latentSplat~\cite{wewer2024latentsplat}, as well as cost volume-based multi-view aggregation in MVSplat~\cite{chen2024mvsplat} and MVSGaussian~\cite{liu2024fast}. Subsequent works, FreeSplat~\cite{wang2024freesplat} and eFreeSplat~\cite{min2024epipolar}, further explore free-view reconstruction and training without epipolar constraints. However, these approaches often encounter limitations when processing scenes with unfamiliar visual content and layouts, as their understanding of the visual world is constrained by training data distribution and hampered by challenges in zero-shot domain generalization. This observation motivates our investigation into leveraging the rich visual priors embedded in state-of-the-art monocular depth foundation models to achieve more robust and generalizable Gaussian reconstruction.

The basis of our approach originates from the observation: Comtemporary monocular depth foundation models (e.g., MiDaS~\cite{ranftl2020towards}, Depth Anything Model~\cite{yang2024depth}), trained on extensive datasets, exhibit exceptional proficiency in predicting monocular depths across diverse visual domains. If a thorough and extensive understanding of the visual world is fundamental to generalizable Gaussian reconstruction, then it should be feasible to develop a widely applicable reconstruction framework using a pretrained monocular depth estimator.
To this end, we present \textbf{MonoSplat}, a novel Gaussian reconstruction framework built upon monocular depth foundation models. The MonoSplat framework comprises two key architectural components: First, a Mono-Multi Feature Adapter that transforms the monocular features from the frozen depth foundation model into multi-view features with cross-view awareness. Second, an Integrated Gaussian Prediction module that synergistically integrates monocular and multi-view features to generate precise Gaussian primitives. Through such an elegant design philosophy, the architecture achieves remarkable efficiency with minimal trainable parameters. Most significantly, MonoSplat's novel integration of monocular depth priors enables unprecedented zero-shot generalization capabilities, establishing new performance standards across diverse real-world scenarios. Our primary contributions are:
\begin{itemize}
    \item We propose a simple yet effective framework to
convert a pre-trained monocular depth foundation model into a Gaussian reconstruction model;
    \item We introduce a Mono-Multi Feature Adapter that converts monocular features from the depth foundation model into multi-view features with cross-view awareness;
    \item We present an Integrated Gaussian Prediction module that effectively combines monocular and multi-view features to produce accurate Gaussian primitives;
    \item MonoSplat, a state-of-the-art, generalizable Gaussian reconstruction model that offers excellent performance across a wide variety of natural images.
\end{itemize}

%% file: sec/2_related.tex
\section{Related work}
\label{sec:related}

\noindent \textbf{3D Reconstruction and View Synthesis}.
The landscape of 3D reconstruction from posed images has been fundamentally transformed by breakthroughs in neural rendering~\citep{tewari2022advances} and neural fields~\citep{mildenhall2020nerf,xie2022neural,sitzmann2020implicit}. Modern approaches construct scene representations through differentiable rendering by optimizing image-space photometric errors. While early methods explored voxel-based structures~\citep{nguyen2018rendernet,sitzmann2019deepvoxels,lombardi2019neural}, the field witnessed a paradigm shift towards neural fields with volume rendering~\citep{mildenhall2020nerf,xie2022neural} as the dominant framework. To address computational inefficiencies inherent in these approaches, researchers investigated discrete data structures~\citep{fridovich2022plenoxels,chen2022tensorf,liu2020neural,muller2022instant}, though the challenge of achieving real-time high-resolution rendering persisted. The advent of 3D Gaussian splatting~\citep{kerbl20233d} introduced efficient rasterization capabilities, yet its reliance on extensive image sets for quality synthesis remained a limitation. Our work pushes the boundaries of this domain by introducing neural architectures capable of estimating 3D Gaussian primitives through single forward passes.

\noindent \textbf{Generalizable Neural Radiance Fields}.
The field of generalizable reconstruction has witnessed remarkable progress since the introduction of Neural Radiance Fields (NeRF)~\citep{mildenhall2020nerf}, enabling advancements in both object-centric~\citep{Chibane_2021_CVPR,Henzler_2021_CVPR,johari2022geonerf,Liu_2022_CVPR,Reizenstein_2021_ICCV,wang2021ibrnet,yu2021pixelnerf} and scene-level reconstruction~\citep{Chibane_2021_CVPR,du2023learning,chen2021mvsnerf,chen2023explicit,xu2024murf}. Following PixelNeRF's~\citep{yu2021pixelnerf} groundbreaking work in image-based radiance field reconstruction using pixel-aligned features, the field has seen continuous innovation through enhanced feed-forward models incorporating sophisticated matching mechanisms~\citep{chen2021mvsnerf,chen2023explicit}, transformer-based architectures~\citep{sajjadi2022scene,du2023learning,miyato2023gta}, and novel volumetric representations~\citep{chen2021mvsnerf,xu2024murf}. While MuRF~\citep{xu2024murf} currently sets the state-of-the-art by leveraging target view frustum volumes with (2+1)D CNN architecture, a fundamental challenge persists across these approaches: the computational burden of per-pixel volume sampling during rendering. The requirement for extensive sampling points along each ray to achieve high-quality reconstruction results in prohibitive computational costs and rendering times, significantly impeding practical deployment.

\begin{figure*}[t!]
  \centering
  \includegraphics[width=0.9\linewidth]{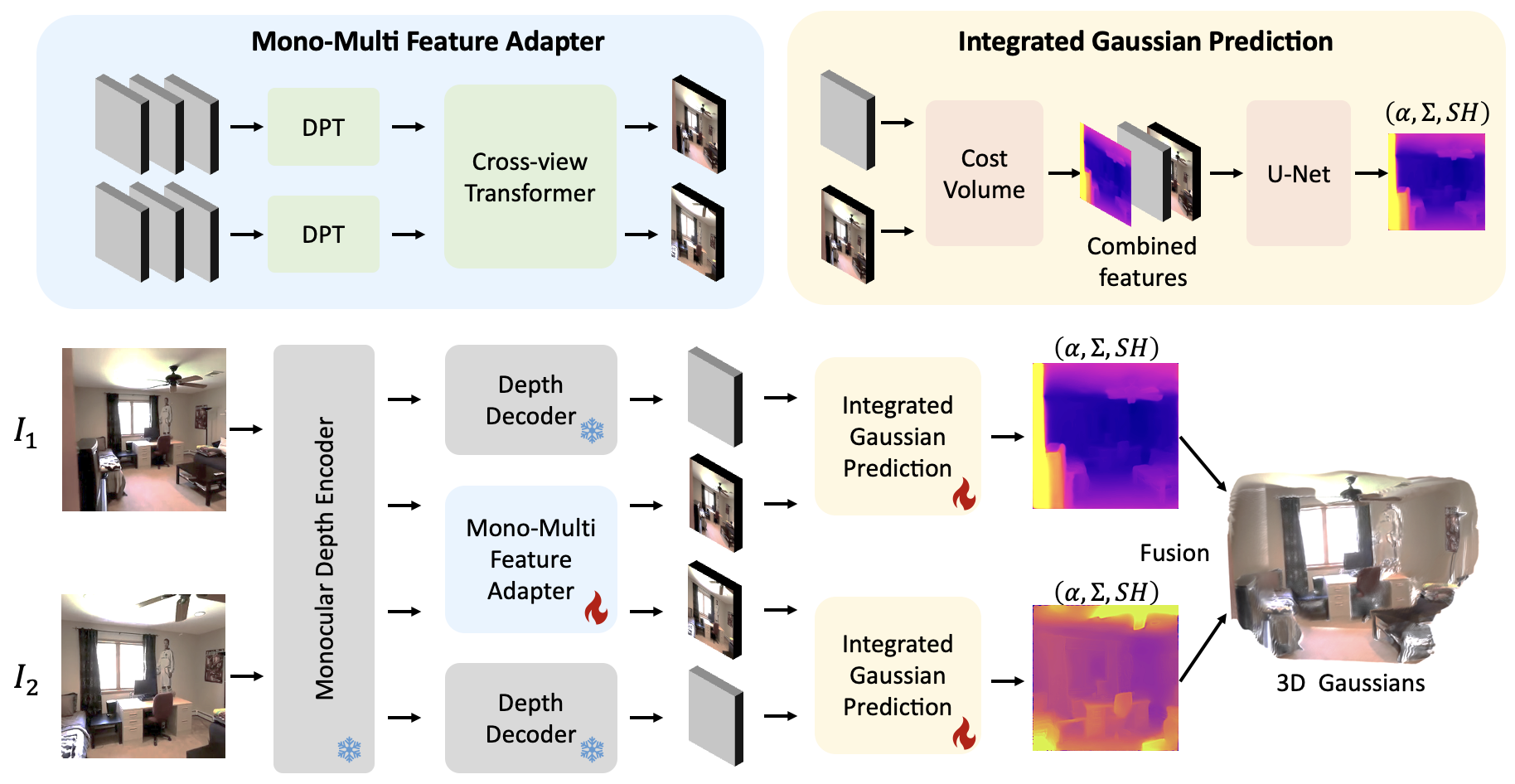}
   \caption{\textbf{Illustration of the proposed MonoSplat framework.} It consists of a Mono-Multi Feature Adapter to convert monocular features to multi-view features with cross-view awareness, and Integrated Gaussian Prediction to integrate monocular features into the Gaussian primitives prediction.}
   \label{fig:framework}
\end{figure*}

\noindent \textbf{Generalizable 3D Gaussian Splatting}.
3D Gaussian Splatting~\citep{kerbl20233d,chen2024survey} introduced an efficient rasterization-based rendering paradigm, sparking significant research interest in feed-forward models~\citep{szymanowicz2023splatter,szymanowicz2024flash3d,charatan2023pixelsplat,zheng2023gps,chen2024mvsplat,liu2024fast,wewer2024latentsplat,zhang2024transplat,tang2024hisplat,fei2024pixelgaussian}. Initial single-view approaches, such as Splatter Image~\citep{szymanowicz2023splatter} and Flash3D~\citep{szymanowicz2024flash3d}, demonstrated promising results in simple scenes but exhibited limitations when handling complex geometries. Then, pixelSplat~\citep{charatan2023pixelsplat} pioneered the use of epipolar transformers and probabilistic depth estimation, while MVSplat~\citep{chen2024mvsplat} and MVSGaussians~\citep{liu2024fast} advanced the field through cost volume-based depth prediction. LatentSplat~\citep{wewer2024latentsplat} further expanded the capabilities with variational Gaussians for view extrapolation. Subsequently, FreeSplat~\cite{wang2024freesplat} addressed reconstruction from extended sequences, and eFreeSplat~\cite{min2024epipolar} introduced a novel architecture that eliminates dependency on epipolar priors. Despite these advances, existing methods have largely overlooked the potential of foundation models' visual priors, limiting their generalizability due to constrained visual knowledge. The concurrent work DepthSplat~\cite{xu2024depthsplat} makes initial strides in leveraging monocular features from Depth Anything Model~\cite{yang2024depth}. However, their dual-branch network architecture introduces redundant model parameters. In contrast, our framework achieves superior efficiency by directly utilizing a frozen depth foundation model and elegantly converting monocular features into multi-view representations, resulting in a more efficient and effective solution.

%% file: sec/3_method.tex
\section{Method}

Given an image sequence $\gI=\{\mI_i\}_{i=1}^N$ ($\mI_i\in\R^{H\times W\times 3}$) and their corresponding camera projection matrices $\gP=\{\mP_i|\mP_i=\mK_i[\mR_i|\vt_i]\}_{i=1}^{N}$, derived from camera intrinsics $\mK_i$, rotation matrices $\mR_i$, and translation vectors $\vt_i$, our objective is to learn a mapping function $f_{\boldsymbol{\theta}}$ that transforms input images to 3D Gaussian primitives:
\begin{equation}
f_{\boldsymbol{\theta}}: \{\mI_i, \mP_i\}_{i=1}^{N} \rightarrow \{\boldsymbol{\mu}_j, \alpha_j, \mathbf{\Sigma}_j, \vc_j\}_{j=1}^{N\times H\times W},
\end{equation}
where $f_{\boldsymbol{\theta}}$ is realized through a feed-forward network with learnable parameters $\boldsymbol{\theta}$, trained on large-scale datasets. Each Gaussian primitive is characterized by its spatial position $\boldsymbol{\mu}_j$, opacity value $\alpha_j$, covariance matrix $\mathbf{\Sigma}_j$, and spherical harmonics-based color coefficients $\vc_j$. To achieve robust and generalizable reconstruction from multiple input views, we propose an architecture that builds upon monocular depth foundation models, as illustrated in Figure \ref{fig:framework}.

The MonoSplat framework consists of two principal components: (1) a Mono-Multi Feature Adapter that transforms monocular features from foundation models into scene-aware multi-view representations (Sec. \ref{sec:mono-multi_feature_adapter}), and (2) an Integrated Gaussian Modeling module that synergistically combines monocular and multi-view features to predict precise Gaussian primitives (Sec. \ref{sec:integrated_gaussian_modeling}). The comprehensive training methodology and optimization strategy are elaborated in Sec. \ref{sec:optimization}.

\subsection{Mono-Multi Feature Adapter}
\label{sec:mono-multi_feature_adapter}
Depth foundation models, with Depth Anything Model~\cite{yang2024depth} as a prominent example, have achieved unprecedented cross-domain generalization through their extensive training on large-scale datasets. However, harnessing these models for generalized Gaussian reconstruction introduces a fundamental challenge: transforming the view-specific monocular representations into geometrically coherent multi-view features that are essential for precise 3D reconstruction. To bridge this gap, we propose a two-stage feature adapter architecture. The first stage consolidates multi-scale encoder features into a unified representation, capturing both fine-grained details and global context. The second stage leverages a multi-view transformer architecture that enables comprehensive cross-view feature exchange, establishing geometric consistency across different viewpoints. Besides, we extract complementary monocular features from the foundation model, providing rich depth priors that serve as a crucial foundation for accurate Gaussian parameter inference.

\subsubsection{Mono-Multi Feature Adaptation}
Our feature extraction and fusion pipeline consists of two main stages: multi-scale feature extraction with unified aggregation, and cross-view geometric reasoning.

In the first stage, for each image $\mI_i$ in the sequence $\gI=\{\mI_i\}_{i=1}^N$ ($\mI_i\in\R^{H\times W\times 3}$), we employ a hierarchical feature extraction strategy to obtain multi-scale representations $\{\mF_i^1, ..., \mF_i^S\}$ from intermediate layers of the monocular depth encoder, where $S$ denotes the number of feature scales. It is worth noting that the depth encoder is deliberately maintained in a frozen state during training, serving dual purposes: preserving the valuable 3D geometric priors acquired through extensive pre-training on large-scale datasets, and substantially reducing the number of trainable parameters. Then, to effectively consolidate these multi-scale features $\{\mF_i^1, ..., \mF_i^S\}$ into a unified representation, we adopt the DPT architecture~\cite{ranftl2021vision}, which progressively fuses the feature hierarchy through:
\begin{equation}
\mF_i = \mathcal{F}_{DPT}({\mF_i^1, ..., \mF_i^S}), \text{  } \mF_i \in \R^{H/4\times W/4\times C},
\end{equation}
\noindent where $C$ represents the feature dimension. Our choice of DPT is motivated by its specialized design for Vision Transformer (ViT) encoders in dense prediction tasks, making it particularly well-suited for integration with depth foundation models. This multi-scale feature fusion is fundamental to our approach, as it enables the aggregated features $\mF_i$ to simultaneously capture fine-grained geometric details and comprehensive scene-level structural information.

In the second stage, we enhance the multi-view geometric understanding through an efficient multi-view Transformer architecture. For view $i$, the cross-view feature refinement can be expressed as:
\begin{equation}
\mF_i^{mv} = \mathcal{F}_{cross}(\mF_i, {\mF_j}_{j\in \mathcal{N}i}), \text{ } \mF_i^{mv}\in \R^{H/4\times W/4\times C_{mv}}
\end{equation}
\noindent where $\mathcal{N}_i$ represents the $M$ nearest neighboring views of view $i$, and $\mathcal{F}_{cross}$ denotes the Transformer operations incorporating both self-attention and cross-attention mechanisms. To achieve computational efficiency without sacrificing effectiveness, we implement the local window attention mechanism introduced in Swin Transformer~\cite{liu2021swin}. For sequences containing more than two views, we introduce an efficient attention computation strategy where each view only performs cross-attention operations with its $M$ nearest neighbors. This approach significantly reduces computational complexity while maintaining the quality of feature interactions across views. The resulting cross-view aware features $\mF_i^{mv}$ effectively encode rich geometric relationships and correspondences across multiple viewpoints. 

\subsubsection{Monocular Feature Extraction}
In addition to the geometry-aware multi-view features $\mF_i^{mv}$, we also leverage the rich monocular priors by extracting features $\mF_i^{mono}$ from the frozen depth foundation model. Rather than utilizing features from the encoder's final layer, we strategically extract features from the decoder's last layer. This design choice is motivated by two considerations: First, the decoder features have undergone comprehensive multi-scale feature aggregation, enabling more precise depth estimation. Second, these features have been refined through the entire decoder hierarchy, incorporating both local details and global context, thus providing more direct and informative depth cues for subsequent Gaussian prediction. This approach harnesses the foundation model's pre-trained knowledge while maintaining computational efficiency through feature reuse.

\subsection{Integrated Gaussian Prediction}
\label{sec:integrated_gaussian_modeling}
Although multi-view features $\mF^{mv}_i$ enable sophisticated cost volume construction for Gaussian parameter estimation, they face inherent limitations in challenging scenarios such as occlusions, texture-less regions, and specular surfaces. To address these fundamental limitations, we propose an innovative approach that integrates rich monocular features $\mF^{mono}_i$ from the foundation model into both the cost volume computation and subsequent parameter prediction stages. This strategic integration allows our framework to leverage the comprehensive geometric priors acquired through large-scale pre-training, enabling robust Gaussian prediction across diverse and challenging conditions.

\subsubsection{Integrated Cost Volume}
Our framework employs a plane-sweep stereo approach to encode cross-view feature-matching information. For each view $i$, we first sample a set of uniform depth values $\{d_m\}_{m=1}^D$ between predefined near and far bounds. The warping process proceeds by back-projecting pixels from the reference view $i$ to 3D points using each sampled depth $d_m$, followed by projecting these 3D points onto neighboring views $j$ through their respective camera matrices. This geometric transformation enables us to warp features from neighboring views $\mF_j^{mv}$ to align with the reference view $i$. We then compute feature correlations $\mC_{j\rightarrow i}$ through dot product operations between the warped features $\{\mF^{mv}_{j\rightarrow i}\}_{m=1}^D$ and reference view features $\mF_i^{mv}$. The final cost volume $\mC_i\in\R^{H/4\times W/4\times D}$ is constructed by aggregating correlation estimates from $M$ neighboring views, enhancing the robustness of our feature matching process.

The key innovation lies in our integration strategy, where we strategically combine both monocular and multi-view features $\mF_i^{mono}$ and $\mF_i^{mv}$ with the initial cost volume $\mC_i$. This integration is particularly powerful as the monocular features, derived from foundation models pretrained on diverse datasets, provide rich geometric priors that complement traditional multi-view matching. These priors are especially valuable in challenging scenarios where multi-view matching often fails. The integrated representation undergoes refinement through a dedicated network $\mathcal{F}_{cost}$:
\begin{equation}
\hat{\mC_i}= \mathcal{F}_{cost}([\mC_i, \mF_{mono}, \mF_{mv}]).
\end{equation}
By leveraging both data-driven priors from monocular features and geometric constraints from multi-view matching, our integrated cost volume achieves more robust and accurate depth estimation across diverse real-world scenes.

\subsubsection{Gaussian Parameter Prediction}
The refined cost volume $\hat{\mC_i}$ undergoes probabilistic processing through a softmax operation, yielding probability distributions $\mP_i\in \R^{H/4\times W/4\times D}$. These probabilities enable the computation of depth maps $\mD_i\in\R^{H/4\times W/4}$ through weighted averaging of depth candidates for each view. While cost volume-based approaches inherently face limitations in resolution and discrete matching characteristics, we address these challenges through a refinement strategy that leverages both geometric and learned priors.

Our refinement process begins by upsampling both the depth maps and features through bilinear interpolation. Crucially, we integrate the interpolated monocular features alongside multi-view features and depth maps through a specialized refinement network:
\begin{equation}
\hat{\mF_i}=\mathcal{F}_{feature}([int(\mD_i), int(\mF^{mono}_i), int(\mF^{mv})]),
\end{equation}
where $int(\cdot)$ represents our interpolation function that upsamples inputs to match the image resolution. This integration of monocular features is particularly beneficial for accurate Gaussian parameter prediction, as it provides rich geometric priors that help resolve ambiguities in-depth estimation and local geometry characterization.

The refined features $\hat{\mF_i}\in\R^{H\times W\times D}$ are then processed by two specialized decoder heads: $\mathcal{F}_{depth}$ for precise inference of Gaussian positions $\boldsymbol{\mu}_i$ and opacities $\alpha_i$, and $\mathcal{F}_{raw}$ for determining Gaussian covariances $\mathbf{\Sigma}_i$ and colors $\vc_i$:
\begin{equation}
\boldsymbol{\mu}_i, \alpha_i = \mathcal{F}_{depth}(\hat{\mF_i}), \text{ } \mathbf{\Sigma}_i, \vc_i = \mathcal{F}_{raw}(\hat{\mF_i}).
\end{equation}

Finally, we aggregate Gaussian attributes from all input images $\{\mI_i\}_{i=1}^N$ following established methodologies~\citep{charatan2023pixelsplat,chen2024mvsplat}. The unified merging operation yields a consolidated Gaussian representation:
\begin{equation}
\mG={\bigcup \{\boldsymbol{\mu}_i, \alpha_i, \mathbf{\Sigma}_i, \vc_i\}_{i=1}^{N}}
\end{equation}
This merged representation, enhanced by the integration of monocular priors throughout our pipeline, captures detailed geometric and appearance information that enables high-fidelity novel view synthesis even in challenging scenarios where traditional multi-view approaches might struggle.

\subsection{Optimization}
\label{sec:optimization}
Our model generates 3D Gaussian primitives for novel view synthesis through a differentiable rendering pipeline. During training, we optimize the network parameters using a combination of reconstruction losses against ground truth RGB images. Specifically, following~\cite{charatan2023pixelsplat,chen2024mvsplat}, we employ a weighted sum of mean squared error ($\mathcal{L}_{mse}$) and perceptual LPIPS~\cite{zhang2018unreasonable} losses:
\begin{equation}
\mathcal{L}= \mathcal{L}_{mse}+ \lambda_{lpips}\mathcal{L}_{lpips}
\end{equation}
\noindent where $\lambda_{lpips}$ balances the contribution of pixel-wise and perceptual supervision.

%% file: sec/4_experiments.tex
\begin{table*}[t]
\centering
\footnotesize

\caption{
\textbf{Quantitative comparisons.} We surpass all baseline methods in terms of PSNR, LPIPS, and SSIM for novel view synthesis on the real-world RealEstate10k~\cite{DBLP:journals/tog/ZhouTFFS18} and ACID~\cite{liu2021infinite} datasets. Furthermore, our method demonstrates relatively low inference costs and reduced memory usage. We highlight \textbf{first-place} results in bold and \underline{second-place} results with underlines in each column.}
\begin{tabular}{l|ccc|ccc|c|c|c}
\toprule
\multirow{2}{*}{\textbf{Method}} & \multicolumn{3}{c|}{\textbf{RealEstate10k~\cite{DBLP:journals/tog/ZhouTFFS18}}} & \multicolumn{3}{c|}{\textbf{ACID~\cite{liu2021infinite}}} & \multirow{2}{*}{\textbf{Time (s) $\downarrow$}} & \multirow{2}{*}{\textbf{Params (M) $\downarrow$}} & \multirow{2}{*}{\textbf{Mem. (GB)$\downarrow$}} \\
& PSNR $\uparrow$ & SSIM $\uparrow$ & LPIPS $\downarrow$ & PSNR $\uparrow$ & SSIM $\uparrow$ & LPIPS $\downarrow$ & & & \\
\midrule
Du et al.\cite{du2023learning} & 24.78 & 0.820 & 0.213 & 26.88 & 0.799 & 0.218 & 1.325 & 125.1 & 19.604 \\
GPNR\cite{suhail2022generalizable} & 24.11 & 0.793 & 0.255 & 25.28 & 0.764 & 0.332 & 13.340 & \underline{9.6} & 19.441 \\
pixelNeRF~\cite{yu2021pixelnerf} & 20.43 & 0.589 & 0.550 & 20.97 & 0.547 & 0.533 & 5.299 & 28.2 & 3.962 \\
MuRF~\cite{xu2024murf} & 26.10 & 0.858 & 0.143 & 28.09 & 0.841 & 0.155 & 0.186 & \textbf{5.3} & N/A \\
\midrule
pixelSplat~\cite{charatan2023pixelsplat} & 26.09 & 0.863 & 0.136 & 28.27 & 0.843 & 0.146 & 0.104 & 125.4 & 3.002 \\
latentSplat~\cite{wewer2024latentsplat} & 23.07 & 0.825 & 0.182 & 24.95 & 0.782 & 0.207 & 0.110 & 187.0 & 3.161 \\
MVSplat~\cite{chen2024mvsplat} & 26.39 & \underline{0.869} & 0.128 & 28.25 & 0.843 & 0.144 & \textbf{0.044} & 12.0 & \underline{0.860} \\
eFreeSplat~\cite{min2024epipolar} & \underline{26.45} & 0.865 & \underline{0.126} & \underline{28.30} & \underline{0.851} & \underline{0.140} & 0.061 & N/A & N/A \\
\textbf{Ours} & \textbf{26.68} & \textbf{0.875} & \textbf{0.123} & \textbf{28.63} & \textbf{0.864} & \textbf{0.138} & \underline{0.051} & 30.3 & \textbf{0.857} \\
\bottomrule
\end{tabular}

\label{tab:comparison}
\end{table*}

\section{Experiments}
We present our experimental evaluation as follows: we first describe our experimental setup (Sec. \ref{sec:exp-setup}), then benchmark our method against state-of-the-art baselines (Sec. \ref{sec:exp-results}), and finally analyze the effectiveness of individual components through ablation studies (Sec. \ref{sec:exp-ablation}).

\subsection{Experimental Setup}
\label{sec:exp-setup}

\noindent \textbf{Datasets.}
Our experiments utilize three diverse datasets: RealEstate10K~\citep{DBLP:journals/tog/ZhouTFFS18}, featuring indoor scenes from YouTube real estate tours (67,477 training, 7,289 test sequences); ACID~\citep{liu2021infinite}, containing aerial landscape footage (11,075 training, 1,972 test sequences); and DTU~\citep{jensen2014large}, comprising object-centric captures under controlled conditions. Following~\cite{charatan2023pixelsplat,chen2024mvsplat}, we evaluate on 16 DTU scenes from 4 viewpoints each. Both RealEstate10K and ACID include precise camera pose annotations, while DTU helps validate cross-domain generalization capability.

\noindent \textbf{Metrics.}
We assess the quality of synthesized views through a comprehensive metric suite. The evaluation framework combines pixel-wise assessment (PSNR), structural analysis (SSIM~\citep{wang2004image}), and perceptual similarity measures (LPIPS~\citep{zhang2018unreasonable}). Alongside these quality metrics, we benchmark model efficiency by measuring inference speed and memory requirements. Following established protocols~\citep{charatan2023pixelsplat,chen2024mvsplat}, we standardize our evaluation at $256\times256$ resolution to enable direct comparisons.

\noindent\textbf{Implementation details.}
Our method is implemented in PyTorch with a custom CUDA backend optimized for 3D Gaussian Splatting operations. The framework's foundation builds upon the ViT-s variant of Depth Anything Model~\cite{yang2024depth}, from which we extract intermediate features at strategic network layers following the official implementation protocol. The architecture incorporates DPT heads comprising four layers with 64-dimensional outputs, complemented by a three-layer cross-view transformer for feature integration. For geometric reasoning, our cost volume construction spans a depth range of $d_{near}=0.5$ to $d_{far}=100.0$, subdivided into $D=128$ discrete planes for comprehensive scene sampling. The training objective balances multiple loss terms through carefully tuned coefficients ($\lambda_{lpips}=0.05$), with optimization conducted over 300,000 iterations using a batch size of 14 on a single A100 GPU. Comprehensive ablation studies exploring different ViT variants and detailed architectural specifications are presented in the supplementary materials.

\subsection{Results}
\label{sec:exp-results}

\noindent \textbf{Baselines.}
We benchmark our approach against several leading methods in scene-level novel view synthesis, spanning three categories: \textbf{i)} Light Field Network approaches, including GPNR~\citep{suhail2022generalizable} and AttnRend~\citep{du2023learning}, \textbf{ii)} NeRF-based solutions represented by pixelNeRF~\citep{yu2021pixelnerf} and MuRF~\cite{xu2024murf}, and \textbf{iii)} state-of-the-art 3D Gaussian Splatting methods, notably pixelSplat~\citep{charatan2023pixelsplat}, latentSplat~\cite{wewer2024latentsplat}, MVSplat~\citep{chen2024mvsplat}, and eFreeSplat~\cite{min2024epipolar}.

\begin{figure*}[t!]
  \centering
  \includegraphics[width=0.9\linewidth]{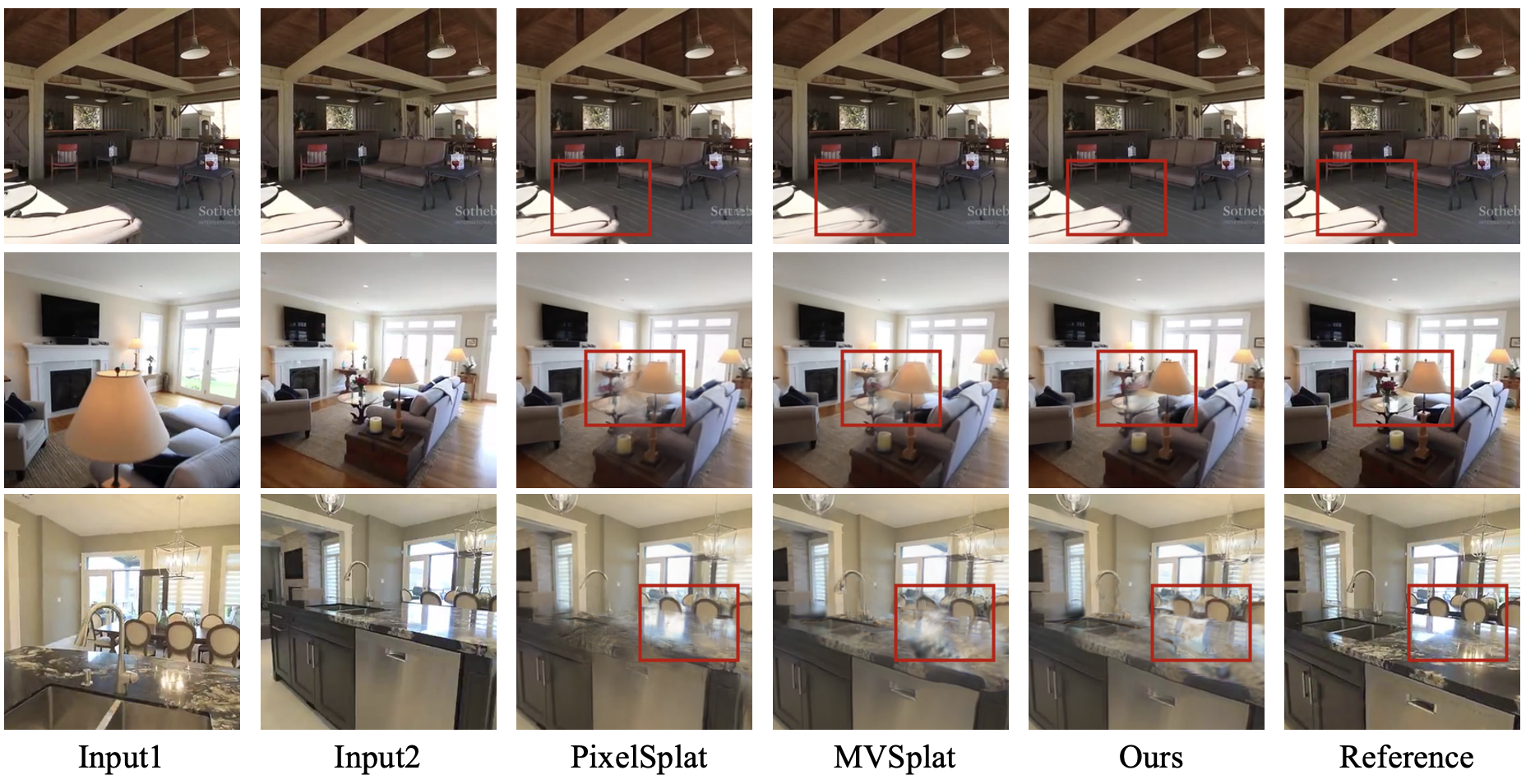}
   \caption{\textbf{Visual comparisons of novel view synthesis on RealEstate10k~\cite{DBLP:journals/tog/ZhouTFFS18}.} Models are trained with a collection of scenes from RealEstate10k, and tested on novel scenes from the test splits. MonoSplat shows superior rendering quality in challenging regions.}
   \label{fig:sota_comparison}
\end{figure*}

\noindent\textbf{Assessing image quality.}
As demonstrated in Table~\ref{tab:comparison}, MonoSplat achieves state-of-the-art performance across all visual quality metrics on both RealEstate10K~\cite{DBLP:journals/tog/ZhouTFFS18} and ACID~\cite{liu2021infinite} benchmarks, attributed to our innovative incorporation of depth foundation models into the Gaussian reconstruction framework. Qualitative comparisons with leading open-sourced models in Figure~\ref{fig:sota_comparison} further substantiate our method's superiority. Notably, MonoSplat demonstrates exceptional performance in challenging scenarios: maintaining fidelity in extreme close-up views (``sofa corner'' in 1st row), preserving geometric consistency under large viewpoint variations (``lamp'' in 2nd row), and accurately rendering regions with substantial illumination changes (``desktop'' in 3rd row). While baseline approaches exhibit visible artifacts in these demanding cases, MonoSplat maintains visual consistency. These results underscore the significance of leveraging pre-trained geometric priors for enhancing the quality and robustness of novel view synthesis across diverse real-world scenarios.

\begin{table}
    \footnotesize
    \centering
    \caption{\textbf{Quantative comparisons of cross-dataset generalization.} We perform zero-shot tests on ACID~\cite{liu2021infinite} and DTU~\cite{jensen2014large} datasets, using models trained solely on RealEstate10K~\cite{DBLP:journals/tog/ZhouTFFS18}. Best and second best results are \textbf{bolded} and \underline{underlined}.}
    \setlength{\tabcolsep}{4.5pt} 
    \begin{tabular}{l|ccc|ccc}
        \toprule
        \multirow{2}{*}{\textbf{Method}} & \multicolumn{3}{c|}{\textbf{Re10k$\rightarrow$DTU}} & \multicolumn{3}{c}{\textbf{Re10k$\rightarrow$ACID}}\\
        & PSNR & SSIM & LPIPS & PSNR & SSIM & LPIPS \\
        \midrule
        pixelSplat~\cite{charatan2023pixelsplat} & 12.89 & 0.382 & 0.560 & 27.64 & 0.830 & 0.160 \\
        MVSplat~\cite{chen2024mvsplat} & \underline{13.94} & \underline{0.473} & \underline{0.385} & \underline{28.15} & \underline{0.841} & \underline{0.147} \\
        \textbf{Ours} & \textbf{15.25} & \textbf{0.605} & \textbf{0.291} & \textbf{28.24} & \textbf{0.848} & \textbf{0.145} \\
        \bottomrule
    \end{tabular}
        \label{tab:cross}
\end{table}

\begin{figure}[t!]
  \centering
  \includegraphics[width=0.95\linewidth]{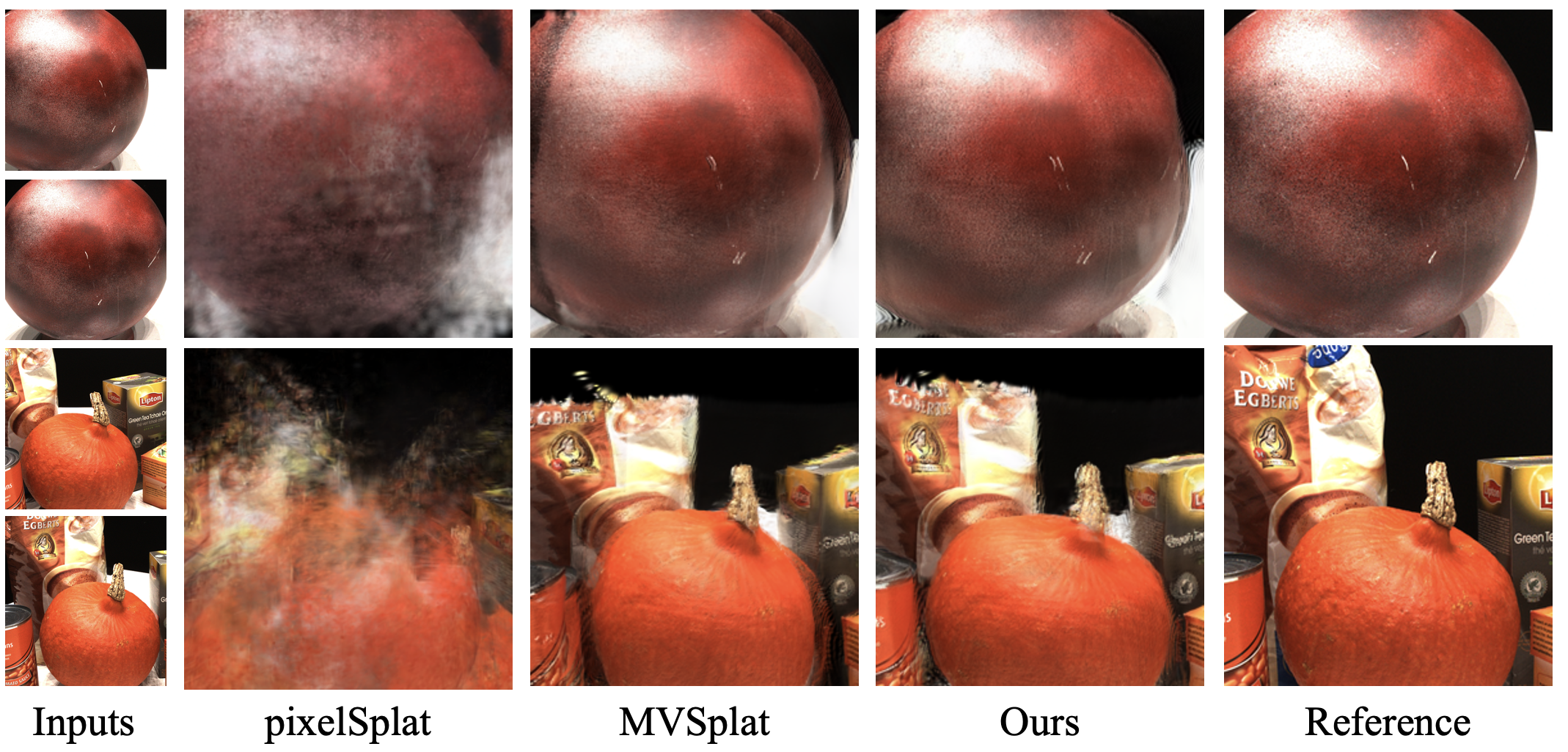}
   \caption{\textbf{Visual comparisons of cross-dataset generalization.} Using a model trained on RealEstate10k~\cite{DBLP:journals/tog/ZhouTFFS18}, we directly render scenes from DTU~\cite{jensen2014large} datasets. MonoSplat shows superior generalization ability compared to previous methods.}
   \label{fig:cross_comparison}
\end{figure}

\noindent\textbf{Assessing cross-dataset generalization.}
Our method demonstrates exceptional generalization capabilities on \emph{out-of-distribution} novel scenes. To rigorously evaluate this advantage, we conduct comprehensive cross-dataset experiments, utilizing models trained exclusively on RealEstate10K indoor scenes for inference on both ACID outdoor scenes and DTU object-centric captures. As evidenced in Figure~\ref{fig:cross_comparison}, MonoSplat consistently produces high-fidelity novel views despite substantial variations in camera distributions and scene characteristics. In comparison, pixelSplat exhibits significant performance degradation, attributable to its feature aggregation strategy's limitations in domain adaptation, while MVSplat demonstrates reduced effectiveness without robust feature injection mechanisms. These qualitative observations are conclusively validated by quantitative metrics presented in Table~\ref{tab:cross}, where MonoSplat achieves substantial improvements in both PSNR and LPIPS metrics across datasets. Notably, the performance disparity between our method and existing approaches becomes increasingly pronounced as the domain gap widens from ACID to DTU datasets, underscoring our method's superior cross-domain generalization capabilities.

\noindent\textbf{Assessing model efficiency.}
As illustrated in Table~\ref{tab:comparison}, MonoSplat demonstrates superior performance across both quality metrics and computational efficiency metrics compared to existing approaches. Our method achieves state-of-the-art image quality while maintaining competitive inference speeds - with an encoding time of 0.051s that closely matches MVSplat's 0.042s. While our total parameter count (30.3M) exceeds MVSplat's (12.0M) due to the incorporation of depth foundation models, MonoSplat actually requires fewer trainable parameters during training (10.3M). This efficiency stems from our architectural design choice to leverage frozen pre-trained depth models, allowing the majority of parameters to remain fixed. Such design not only reduces memory requirements below those of MVSplat but also enables efficient fine-tuning for downstream tasks with minimal computational and memory overhead.

\begin{figure*}[t!]
  \centering
  \includegraphics[width=0.95\linewidth]{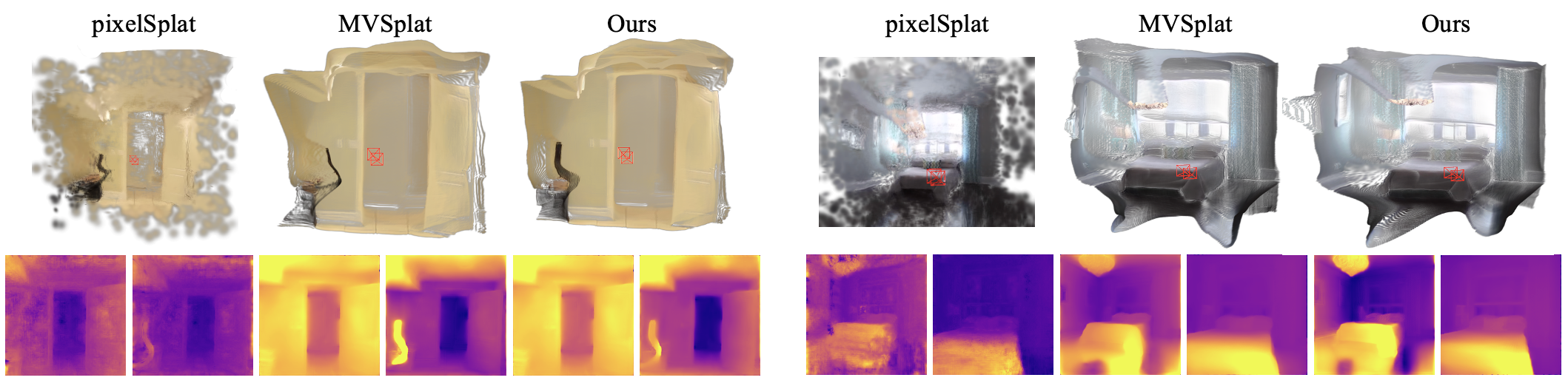}
   \caption{\textbf{Visual comparisons of predicted 3D Gaussians (top) and depth maps (bottom).} We examine the quality of geometric reconstruction by presenting comparative visualizations of 3D Gaussian primitives generated by pixelSplat, MVSplat, and our MonoSplat, alongside depth maps rendered from two reference viewpoints.}
   \label{fig:geo_comparison}
\end{figure*}

\noindent\textbf{Assessing geometry reconstruction.}
Figure~\ref{fig:geo_comparison} showcases MonoSplat's superior capability in generating high-quality 3D Gaussian primitives compared to state-of-the-art methods pixelSplat~\cite{charatan2023pixelsplat} and MVSplat~\cite{chen2024mvsplat}, achieving exceptional geometry reconstruction with purely photometric supervision. While pixelSplat delivers acceptable 2D rendering quality, its 3D reconstruction exhibits significant artifacts, particularly floating Gaussians that fail to adhere to the true geometric structure. Similarly, MVSplat demonstrates limitations in depth estimation accuracy, especially in regions with complex geometric details. In contrast, MonoSplat produces notably more precise and coherent 3D Gaussians, leveraging our effective integration of monocular features to provide robust geometric priors.. 

\subsection{Ablations}
\label{sec:exp-ablation}

\begin{figure}[t!]
  \centering
  \includegraphics[width=0.95\linewidth]{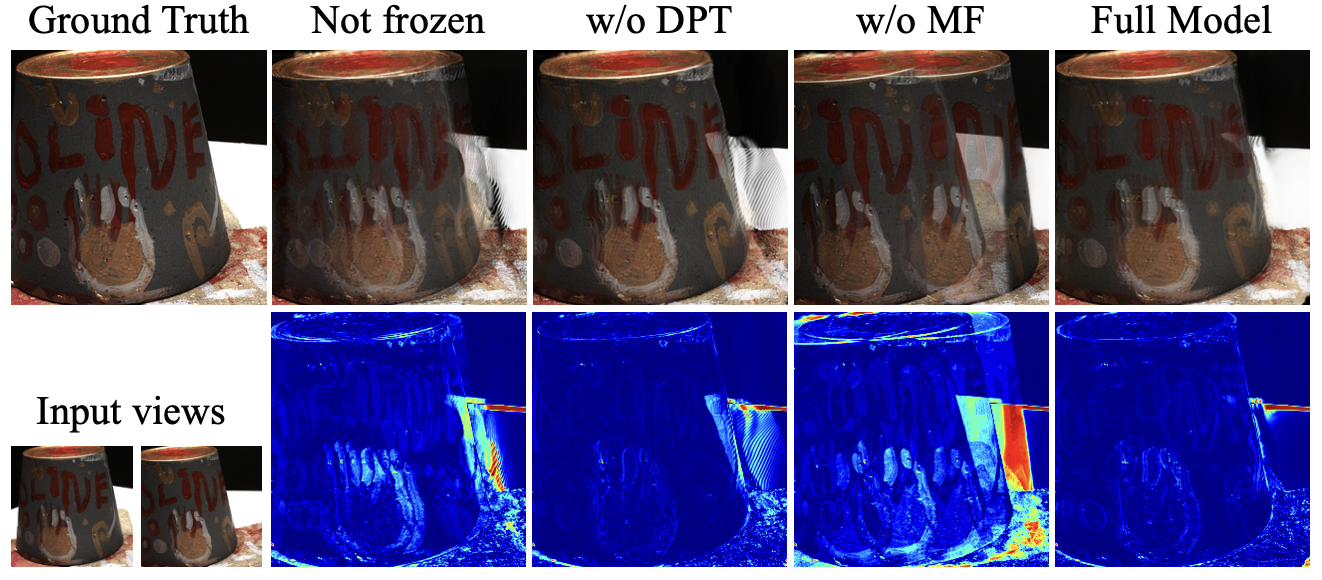}
   \caption{\textbf{Visual comparisons on Re10k$\rightarrow$ DTU with different designs.} Models are trained on RealEstate10k~\cite{DBLP:journals/tog/ZhouTFFS18} and directly tested on DTU~\cite{DBLP:journals/tog/ZhouTFFS18}. Color-coded error distributions are illustrated with the pixel-wise differences between synthesized outputs and ground truth images. }
   \label{fig:ablation}
   \vspace{-0.2cm}
\end{figure}

We conduct thorough ablations on RealEstate10K~\cite{DBLP:journals/tog/ZhouTFFS18} to analyze MonoSplat, where all models are trained for 20, 000 iterations with a batch size of 14. Results are shown in~\cref{tab:ablation} and~\ref{fig:ablation}, and we discuss them in detail next.

\noindent \textbf{Importance of frozening depth foundation model.}
We hypothesize that the strong geometric priors from the depth foundation model are crucial for superior generalization in Gaussian reconstruction. To validate this, we compare our frozen pre-trained foundation model against a variant ``Not frozen'' where the backbone is fine-tuned during training. As shown in Table~\ref{tab:ablation}, allowing backbone updates leads to notable performance degradation, particularly in generalization capability (2.61dB decrement). This occurs because fine-tuning on limited scene data causes the model to overfit and lose the rich geometric priors. Keeping the foundation model frozen preserves these valuable priors for better cross-scenario reconstruction, as demonstrated by the qualitative results in Figure \ref{fig:ablation}, where the fine-tuned model shows more errors on out-of-domain data.

\noindent \textbf{Importance of DPT.}
The DPT component plays a crucial role in aggregating multi-scale features from the frozen encoder, which is essential for subsequent feature matching. When we remove DPT and directly utilize features from the final encoder layer in our experiments, we observe a dramatic decline in reconstruction quality, as shown in Table~\ref{tab:ablation} and Figure~\ref{fig:ablation}. This performance degradation can be attributed to the limitation of single-scale features in supporting effective feature matching within the cost volume. These results validate the efficacy of incorporating DPT for robust feature aggregation.

\noindent \textbf{Importance of cross-view aggregation.}
Cross-view aggregation enables monocular features to incorporate information from other viewpoints, establishing accurate pixel correspondences. This transformation is critical for subsequent cost volume matching, as the cost volume relies on pixel correspondences to reason about depth. To validate its importance, we design a variant ``w/o cross aggregation'' by removing the cross-view aggregation module. As shown in Table~\ref{tab:ablation}, this leads to degraded reconstruction performance of 0.98 dB decrement, primarily due to the cost volume's inability to establish reliable correspondences.

\noindent \textbf{Importance of monocular feature integration.}
As described in Sec. \ref{sec:integrated_gaussian_modeling}, we integrate monocular features into the cost volume and also the feature refinement network. To validate its effectiveness, we design three variants with ``w/o MF'' that removes monocular features, ``MF in cost volume'' and ``MF in refinement'' that only integrates monocular features during the cost volume construction and feature refinement stages, respectively. As shown in Table~\ref{tab:ablation} and Figure~\ref{fig:ablation}, removing monocular features entirely leads to extreme generalization performance drop, indicating that integrating monocular features is crucial to the generalizability. In addition, integrating monocular features into only one stage can also cause degraded reconstruction performance. These results demonstrate not only the effectiveness of monocular features but also the importance of incorporating them throughout the pipeline to maximize their utility.

\begin{table}
    \footnotesize
    \centering
    \caption{\textbf{Ablation study.} We perform ablation studies using models trained solely on Re10K~\cite{DBLP:journals/tog/ZhouTFFS18} with 200k iterations, and test on the in-domain test set of Re10k and out-domain test set of DTU~\cite{jensen2014large}. Best and second best results are \textbf{bolded} and \underline{underlined}.}
    \setlength{\tabcolsep}{4.5pt} 
    \begin{tabular}{l|cc|cc}
        \toprule
        \multirow{2}{*}{\textbf{Method}} & \multicolumn{2}{c|}{\textbf{Re10k}} & \multicolumn{2}{c}{\textbf{Re10k$\rightarrow$DTU}}\\
        & PSNR $\uparrow$ & SSIM$\uparrow$ & PSNR$\uparrow$ & SSIM$\uparrow$ \\
        \midrule
        Not frozen & 25.87 & 0.860 & 12.63 & 0.328 \\
        \midrule
        w/o DPT & 25.25 & 0.846 & 14.69 & 0.480 \\
        w/o cross aggregation& 25.48 & 0.858 & 10.28 & 0.267 \\
        \midrule
        w/o MF & \underline{26.33} & \textbf{0.870} & 10.33 & 0.268 \\
        MF in cost volume & 26.20 & \underline{0.867} & \underline{15.06} & 0.466 \\
        MF in refinement& 26.06 & 0.863 & 14.93 & \underline{0.527} \\
        \midrule
        Full model & \textbf{26.50} & \textbf{0.870} & \textbf{15.24} & \textbf{0.604} \\
        \bottomrule
    \end{tabular}
        \label{tab:ablation}
        \vspace{-0.1cm}
\end{table}

%% file: sec/5_conclusion.tex
\section{Conclusion}
\label{sec:conclusion}
In this paper, we present MonoSplat, a framework that improves 3D Gaussian reconstruction by leveraging pretrained monocular depth models. Our approach transforms monocular priors into accurate Gaussian primitives while maintaining cross-view consistency. MonoSplat achieves state-of-the-art performance on diverse real-world scenes and demonstrates strong zero-shot generalization with minimal computational overhead, highlighting the value of incorporating geometric priors from foundation models.

%% file: sec/6_acknowledgement.tex
\section{Acknowledgment}

This work was supported by Hong Kong Research Grants Council (RGC) General Research Fund 11211221 and 14204321, and conclusions contained herein reflect the opinions and
conclusions of its authors and no other entity.

%% file: sec/X_suppl.tex
\clearpage
\setcounter{page}{1}
\maketitlesupplementary

\appendix

\section{More Details}
\label{sec:details}

\subsection{Data}
During training, we employ our custom data loaders for all methods and progressively increase the spacing between reference views. Specifically, we implement a linear increase in view distances over the first 150,000 training steps: the minimum distance between reference views increases from 25 to 45, while the maximum distance expands from 45 to 192. To ensure a fair comparison with previous works~\cite{charatan2023pixelsplat,chen2024mvsplat}, input images are resized to a resolution of 256×256. Our pre-processing filters out invalid images, including those with misaligned sizes and images where the maximum field of view exceeds 100 degrees.

\subsection{Model}
For the frozen depth encoder, we adopt three variants of Depth Anything V2~\cite{yang2024depth}, which are based on ViT-S, ViT-B, and ViT-L, respectively. For these variants, we extract features of intermediate layers of [2, 5, 8, 11], [2, 5, 8, 11], and [4, 11, 17, 23], respectively, and feed these features to the following DPT decoder and the original depth decoder. For DPT, we set the final output dimension as 64 to balance the efficiency and effectiveness. Following DPT, the multi-view transformer adds position encoding for different views and outputs features of dimension 64. 

For the cost volume construction, we use 128 planes and calculate the 2D cost volume, similar to \cite{chen2024mvsplat}, followed by the UNet-style refinement. The cost volume UNet uses a base feature dimension of 128, which was empirically found to balance model capacity and efficiency well. We maintain consistent channel dimensions across three downsampling stages using multipliers [1,1,1]. Self-attention is applied at 1/4 resolution to enhance feature correlation.

After obtaining the predicted depths, we combine them with features for further Gaussian parameter prediction, achieved by a depth UNet. The depth UNet employs a base feature dimension of 32, with channel multipliers [1,1,1,1,1] across five downsampling stages. Attention mechanisms are incorporated at resolution 1/16 to capture long-range dependencies. Finally, we constrain the Gaussian scale within the range [0.5, 15.0]. The minimum scale of 0.5 ensures sufficient detail capture at fine levels. The maximum scale of 15.0 prevents overly large Gaussians while allowing coverage of broader regions. We use spherical harmonics of degree 4 to represent view-dependent appearance.

\subsection{Training}
Our default model training is conducted on a single A100 GPU with a batch size of 14. Each batch comprises one training scene consisting of two input views and four target views. Following pixelSplat~\cite{charatan2023pixelsplat} and MVSplat~\cite{chen2024mvsplat}, we progressively increase the frame distance between input views throughout the training process. The near and far depth planes are empirically set to 0.5 and 100 for both RealEstate10K and ACID datasets. For the DTU dataset, we utilize the depth bounds of 2.125 and 4.525.

\section{More Experimental Analysis}
All experiments in this section follow the same settings as in Sec. \ref{sec:exp-setup} unless otherwise specified, which are trained and tested on RealEstate10K~\cite{DBLP:journals/tog/ZhouTFFS18}. To investigate our hypothesis regarding the advantages of monocular depth foundation model features for generalizable Gaussian reconstruction, we conducted experiments with various frozen backbones, including DINOv2 (used in pixelSplat~\cite{charatan2023pixelsplat}) and UniMatch (employed in MVSplat~\cite{chen2024mvsplat}). The results, as presented in the Table \ref{tab:ablation_backbone}, show substantial deterioration in both in-domain and out-of-domain generalization performance, validating the essential role of depth foundation models. Furthermore, our exploration of different Depth Anything v2 variants revealed a positive correlation between model size and performance, with larger models achieving better reconstruction quality and generalization capabilities.

\begin{table}
    \small
    \centering
    \caption{\textbf{Ablation on the backbone.} We perform ablation studies using different backbones trained solely on Re10K~\cite{DBLP:journals/tog/ZhouTFFS18} with 200k iterations, and test on the in-domain test set of Re10k and out-domain test set of DTU~\cite{jensen2014large}. }
    \begin{tabular}{l|cc|cc}
        \toprule
        \multirow{2}{*}{\textbf{Method}} & \multicolumn{2}{c|}{\textbf{Re10k}} & \multicolumn{2}{c}{\textbf{Re10k$\rightarrow$DTU}}\\
        & PSNR $\uparrow$ & SSIM$\uparrow$ & PSNR$\uparrow$ & SSIM$\uparrow$ \\
        \midrule
        DINOv2 & 26.14 & 0.872 & 13.45 & 0.342  \\
        UniMatch & 26.03 & 0.868 & 14.92 & 0.465 \\
        \midrule
        DAMv2-S & 26.50 & 0.870 & 15.24 & 0.604 \\
        DAMv2-B & 26.83 & 0.875 & 15.62 & 0.620 \\
        DAMv2-L & 27.12 & 0.878 & 15.95 & 0.608 \\
        \bottomrule
    \end{tabular}
        \label{tab:ablation_backbone}
\end{table}

\section{More Visual Comparisons}
\begin{figure*}[h]
  \centering
  \includegraphics[width=0.95\linewidth]{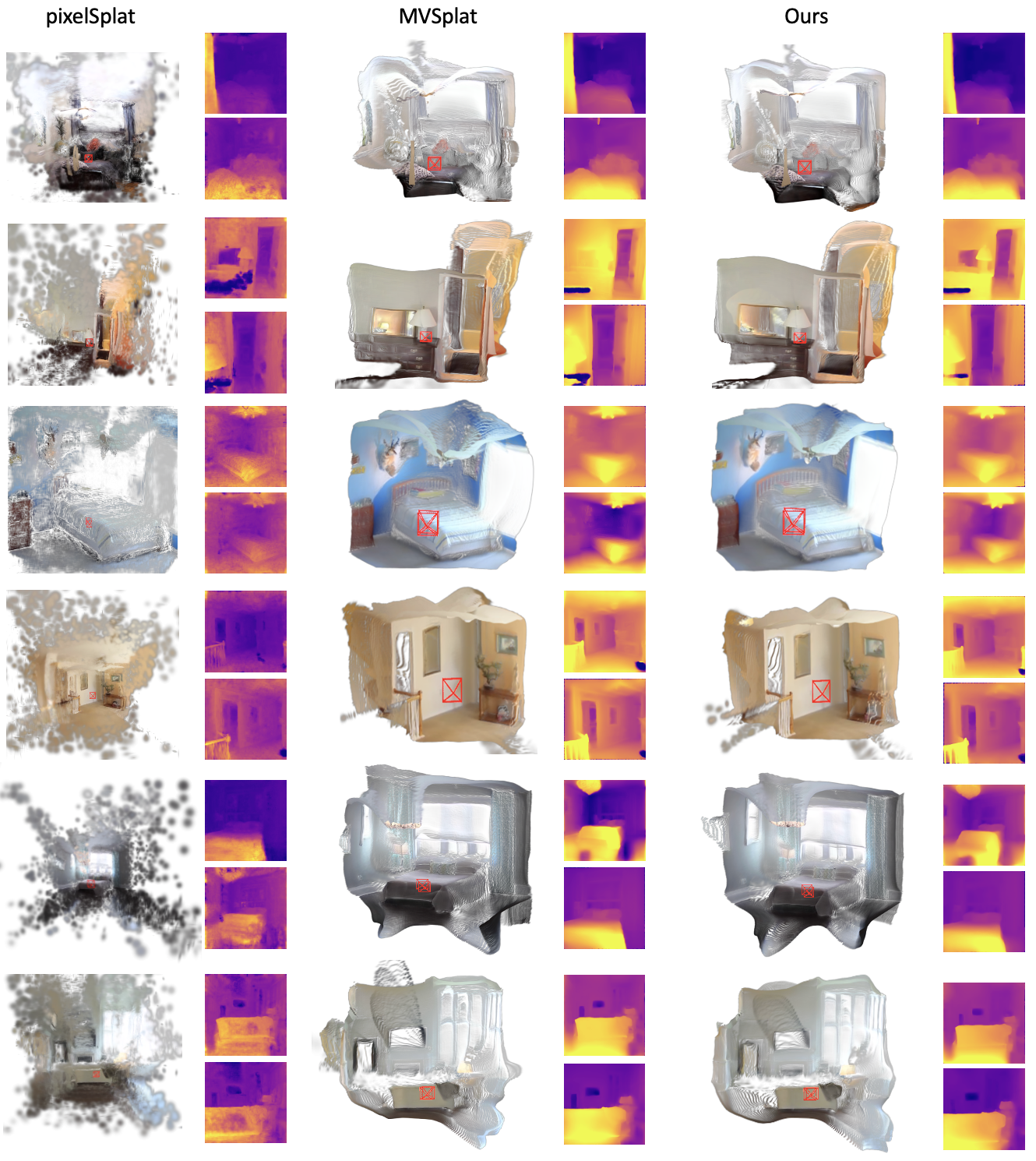}
   \caption{\textbf{Additional visual comparisons of geometry reconstruction on RealEstate10k~\cite{DBLP:journals/tog/ZhouTFFS18}.} All models were trained on a diverse collection of RealEstate10k scenes and evaluated on previously unseen scenes from the test split. For reference, we provide two rendered depth maps from the input views. Our method demonstrates superior reconstruction quality across different viewpoints and scene structures. }
   \label{fig:suppl_geometry_comparison}
\end{figure*}

\begin{figure*}[h]
  \centering
  \includegraphics[width=0.95\linewidth]{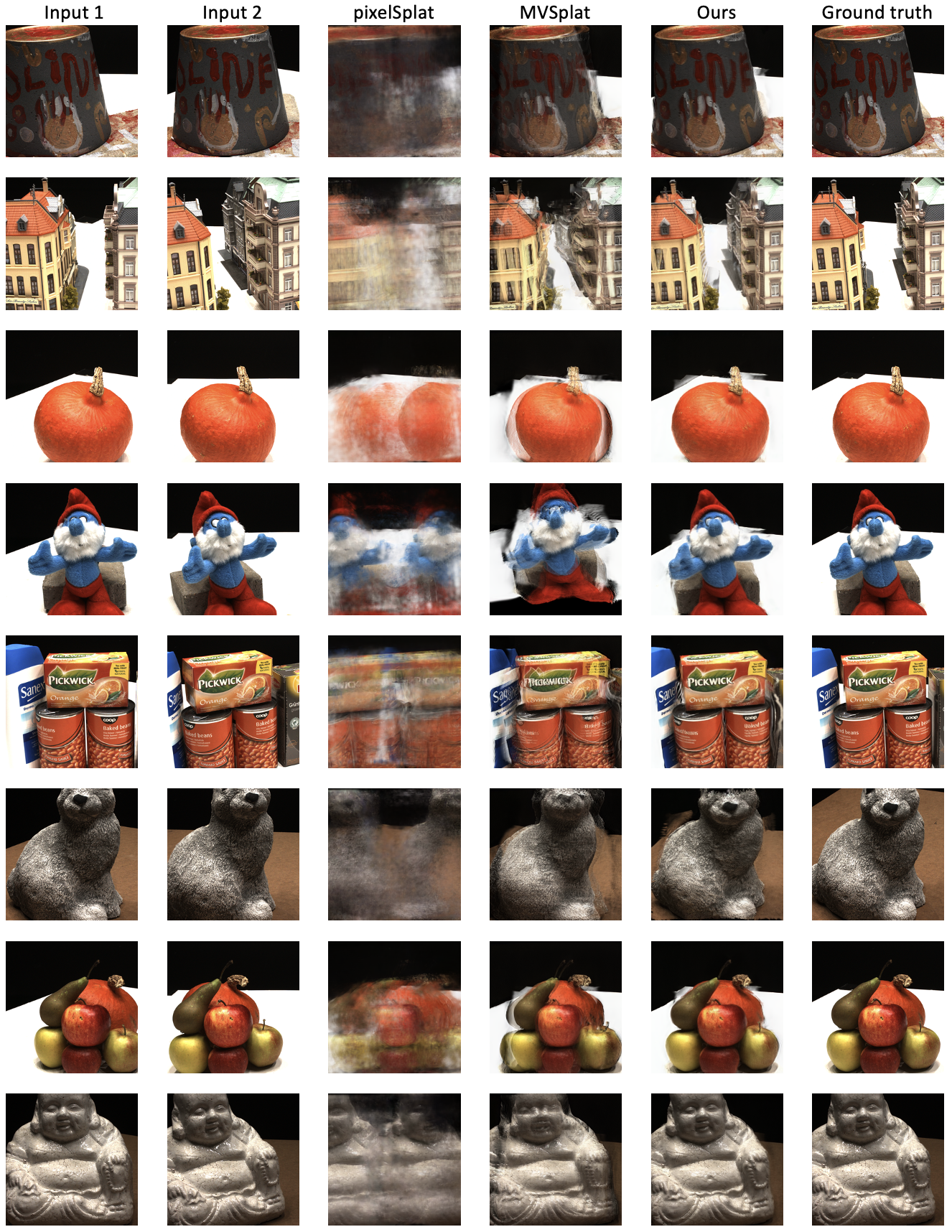}
   \caption{\textbf{More visual comparisons of cross-dataset generalization from RealEstate10k~\cite{DBLP:journals/tog/ZhouTFFS18} to DTU~\cite{DBLP:journals/tog/ZhouTFFS18}.} Models are trained solely on RealEstate10k, and tested on novel scenes from DTU. Our method shows superior rendering quality compared to previous methods. }
   \label{fig:suppl_cross_comparison}
\end{figure*}

In this section, we provide more visual comparisons of geometry reconstruction in Figure ~\ref{fig:suppl_geometry_comparison} and cross-dataset generalization results in Figure~\ref{fig:suppl_cross_comparison}.

